\documentclass[10pt,letterpaper]{article}
\usepackage[letterpaper,top=0.85in,bottom=0.85in,left=1in,right=1in]{geometry}

\usepackage[utf8]{inputenc}
\usepackage{float}      
\usepackage{booktabs}   
\usepackage{adjustbox} 
\usepackage{graphicx}
\usepackage{subcaption}
\usepackage{cite}
\usepackage{seqsplit}
\usepackage{tabularx}
\usepackage{array}
\usepackage{multirow}
\usepackage{booktabs}
\newcolumntype{Y}{>{\centering\arraybackslash}X} 
\newcolumntype{L}{>{\raggedright\arraybackslash}X}
\newcolumntype{C}[1]{>{\centering\arraybackslash}p{#1}}
\newcolumntype{T}[1]{>{\raggedright\arraybackslash}p{#1}} 
\newcommand{\nw}[1]{\mbox{#1}}
\usepackage{nameref,hyperref}
\usepackage{pgfplots}
\pgfplotsset{compat=1.18}

\usepackage{microtype}
\DisableLigatures[f]{encoding = *, family = * }

\usepackage{changepage}

\usepackage[aboveskip=1pt,labelfont=bf,labelsep=period,singlelinecheck=off]{caption}
\makeatletter
\renewcommand{\@biblabel}[1]{\quad#1.}
\makeatother

\usepackage{lastpage,fancyhdr,graphicx}
\usepackage{epstopdf}
\usepackage{color}
\usepackage{makecell}
\usepackage{xcolor}

\usepackage{algorithm}
\usepackage{algpseudocode} 
\usepackage{amsmath,amssymb}
\usepackage{tikz}
\usetikzlibrary{arrows.meta,positioning,shapes.geometric,fit,calc}

\definecolor{Gray}{gray}{.25}

\usepackage{graphicx}

\usepackage{sidecap}

\usepackage{wrapfig}
\usepackage[pscoord]{eso-pic}
\usepackage[fulladjust]{marginnote}

\begin{document}
\vspace*{0.35in}

\begin{flushleft}
{\Large
\textbf\newline{VFA: Relieving Vector Operations in Flash Attention with Global Maximum Pre-computation}
}
\newline
\\
Yupeng Sun\textsuperscript{1,*},
Yanzhao Li\textsuperscript{1,*},
Zhiqiang Zou\textsuperscript{1},
Bai Du\textsuperscript{1},
Zhiyuan Zhang\textsuperscript{1},
Hui Dong\textsuperscript{1},
Gaoyige Fan\textsuperscript{1},
Hui Wang\textsuperscript{1}
\\
\bigskip
\bf{1} Huawei Technologies
\\
\bigskip
* {sunyupeng7, liyanzhao}@huawei.com

\end{flushleft}

\section*{Abstract}
FlashAttention-style online softmax enables exact attention computation with linear memory by streaming score tiles through on-chip memory and maintaining a running maximum/normalizer \cite{dao2022flashattention, dao2023flashattention}.
However, as attention kernels approach peak tensor-core/cube-core throughput on modern accelerators, the non-matmul components of online softmax---in particular, per-tile \texttt{rowmax}/\texttt{rowsum} reductions and the rescale chain---can become \emph{vector/SIMD limited} and dominate end-to-end latency.
This paper revisits FlashAttention and proposes \emph{Vector Relieved Flash Attention} (VFA), a hardware-friendly modification that  reduces the frequency of \texttt{rowmax}-driven updates of the running maximum while preserving the online-softmax accumulation structure. VFA initializes the running maximum using a cheap approximation based on key-block representations, reorders the key-block traversal to prioritize empirically high-impact regions (sink and local blocks), and freezes the running maximum for the remaining blocks to avoid repeated reductions and rescale operations.
We further show that VFA composes naturally with online block-sparse skipping methods such as BLASST \cite{yuan2025blasst}, yielding \emph{Vector Relieved Sparse Attention} (VSA) that combines fewer processed blocks with cheaper per-block load overhead. It's noteworthy that VFA/VSA entirely bypasses the rescale operation in the update stage, an operation which is conditionally executed in FA4.0.
Extensive evaluations on representative benchmarks (e.g., MMLU, MATH500) and numerical analyses of attention statistics validate the design choices: (i) sink+local reordering is supported by early stabilization of the running maximum, (ii) naive block summaries of $Q/K$ are insufficient due to intra-block heterogeneity, and (iii) $m$-initialization is necessary to cover cases where maxima occur in middle blocks.
Overall, VFA/VSA provide a practical path to improving attention efficiency in regimes where online-softmax reductions become a primary kernel bottleneck, without degrading the model performance. 
Compared to the baseline C16V32, the C8V32, C4V32 and C4V16 versions can achieve about 2x speedup based on the latest architecture and they have all reached the vector bottleneck. As the evolution of new architecture,  C4V16 will deliver a 6x speedup by increasing the exponent capacity.


\section{Introduction}
\label{sec:intro}

Transformer attention is a core primitive for modern foundation models across language, vision, and multimodal settings \cite{vaswani2017attention, radford2019language, hatamizadeh2023fastervit}.
Yet, scaled dot-product attention is notoriously expensive for long sequences due to its quadratic complexity and the associated memory traffic for softmax and $PV$ \cite{vaswani2017attention}.
This has motivated substantial work on efficient attention mechanisms, encompassing two main directions: efficient kernel design and low-complexity algorithm design. Low complexity algorithm design primarily includes linear approximations and sparse computation patterns \cite{sun2025efficient}.
FlashAttention is a typical representative of efficient kernel design methods and has emerged as a widely adopted solution for exact attention: it avoids materializing the full attention matrix in HBM by tiling $Q/K/V$ into on-chip memory and updating softmax statistics online, achieving linear memory in sequence length while retaining numerical stability \cite{dao2022flashattention}.

The FlashAttention line of work further demonstrates that efficiency is not purely an algorithmic issue but also a kernel-mapping problem.
FlashAttention-2 improves performance via better parallelism and work partitioning, increasing occupancy and reducing shared-memory communication \cite{dao2023flashattention}.
FlashAttention-3 exploits hardware features such as asynchronous data movement and low-precision pathways (e.g., FP8) to further improve pipeline efficiency on modern GPUs \cite{shah2024flashattention}.
FlashAttention-4 further reduces softmax-side overhead by approximating \texttt{exp2} with low-latency linear instructions and conditionally eliding the output rescale on $\tilde{O}$ when it is numerically safe, thereby shrinking the non-matmul critical path \cite{dao_flashattention_github}.
These advances underline a key trend: once tiled matmuls are highly optimized, attention performance increasingly hinges on the remaining non-matmul components of online softmax, including per-tile reductions (\texttt{rowmax}/\texttt{rowsum}) and the rescale chain used to maintain numerical stability.

\paragraph{Motivation: when online-softmax becomes vector limited.}
To achieve higher throughput and lower latency, models are evolving toward lower precision such as MXFP4\cite{tseng2025training}, NVFP4\cite{abecassis2025pretraining} and HIF4\cite{luo2026hifloat4}.  
However, these lower precisions are only applied in GEMM and vector operations are still in high precision, which bring  higher load imbalance for fused kernels. 
In many practical deployments, we observe that the cost of online-softmax statistic updates can become vector/SIMD limited and disproportionately impacts end-to-end latency, especially when matmul pipelines already run near peak utilization.
This motivates a complementary research question to prior FlashAttention work: \emph{can we reduce the frequency of per-tile reductions and rescale operations, while keeping the same online-softmax accumulation structure and kernel compatibility?}

\paragraph{Our approach.}
We propose VFA, a hardware-friendly variant of FlashAttention-style online softmax.
VFA modifies the standard per-tile update schedule in three steps.
First, it performs a fast \emph{$m$-initialization} that approximates the per-row running maximum using cheap dot products between $Q_i$ and precomputed key-block representations.
Second, it reorders the traversal of key blocks to prioritize an initial ``sink'' block and the local block aligned with the query position, motivated by empirical evidence that attention mass and the eventual running-max stabilization concentrate in these regions.
Third, it freezes the running maximum for the remaining blocks, skipping per-block \texttt{rowmax} and the rescale chain, while still accumulating $\exp(S_{ij}-m_i)$ and $PV$ using the same state variables and stabilization form.
Importantly, VFA targets statistic-computation overhead rather than skipping attention blocks outright.

\paragraph{Composition with FlashAttention-4.}
VFA naturally skip all the rescale operations in attention computation, avoiding online conditional branch in FA4 \cite{zadouri2026flashattention}.

\paragraph{Composition with sparse attention.}
VFA is orthogonal to sparse attention methods that reduce the number of processed blocks.
In particular, BLASST introduces an online-statistics-driven rule that skips blocks whose local maximum is sufficiently below the running maximum, requiring minimal changes to FlashAttention-style kernels \cite{yuan2025blasst}.
We show that VFA composes naturally with BLASST, yielding VSA that combines two multiplicative gains: fewer blocks processed (sparsification) and cheaper per-block load overhead (vector-relieved online softmax).

\paragraph{Empirical analyses supporting the design.}
Beyond end-to-end accuracy evaluations, we provide numerical analyses of attention statistics to justify each component.
We show that the running maximum often stabilizes early at sink/local blocks, explaining why reordering is effective.
We also demonstrate that naive block representations of $Q/K$ are ineffective due to moderate intra-block directional similarity and large norm variability, and that $m$-initialization is necessary because the true block-maximum can occur in middle blocks that would otherwise be missed under max-freezing.

\paragraph{Contributions.}
In summary, this work makes the following contributions:
\begin{itemize}
  \item We identify and study a kernel-level bottleneck in FlashAttention-style online softmax: per-tile reductions and rescale operations can become vector/SIMD limited in regimes where matmul is already efficient.
  \item We propose VFA, a hardware-friendly online-softmax variant that uses $m$-initialization, sink+local reordering, and max-freezing to reduce the frequency of \texttt{rowmax}-driven updates while preserving the standard accumulation structure.
  \item We present VSA by composing VFA with BLASST-style block skipping \cite{yuan2025blasst}, illustrating that vector-relieved load overhead and sparse skipping are complementary, and suggesting that VFA can serve as a reusable building block in future attention kernel evolution.
  \item We provide numerical analyses of attention statistics (running-max stabilization, block similarity, norm variability, and block-max peak locations) that explain when and why each design choice is effective.
  \item We evaluate the performance of VFA based on the compute capability of the latest architecture.
\end{itemize}

\section{Related Works}

\subsection{Vanilla Attention}
Transformer has become the de facto backbone for modern foundation models across language, vision, and multimodal applications, making attention efficiency a first-order concern in both training and inference \cite{radford2019language, hatamizadeh2023fastervit}.
Transformer adopts scaled dot-product attention, which computes
$S = QK^{\top} / \sqrt{d}$, applies a row-wise softmax $P=\mathrm{softmax}(S)$, and produces the output $O = PV$.
Although conceptually simple, vanilla attention is both compute- and memory-intensive for long sequences:
forming $S\in\mathbb{R}^{N_q\times N_k}$ incurs $O(N_qN_kd)$ FLOPs, while explicitly materializing $S$ and $P$ requires
$O(N_qN_k)$ memory and substantial HBM traffic due to multiple reads/writes across the softmax and $PV$ stages \cite{vaswani2017attention}.
In practice, the softmax stabilization (e.g., subtracting the row maximum) and the normalization/reduction steps
(\texttt{rowmax}, \texttt{rowsum}, and rescaling) further introduce bandwidth-heavy operations and synchronization,
making the end-to-end kernel increasingly memory-bound as sequence length grows.
These characteristics motivate IO-aware attention formulations that avoid storing the full attention matrix and
instead stream blocks through on-chip memory while updating softmax statistics online.

\subsection{Efficient Attention Mechanisms}
\label{subsec:sparse_attention}

\paragraph{Flash Attention}
FlashAttention (FA) establishes an IO-aware formulation of exact attention by explicitly accounting for the GPU memory hierarchy \cite{dao2022flashattention}.
Instead of materializing the full attention matrix in HBM, FA tiles the computation so that blocks of $Q$, $K$, and $V$ are staged in on-chip SRAM, and the softmax statistics are updated online using a running maximum and normalizer \cite{milakov2018online}.
This design reduces HBM reads/writes and achieves linear memory in sequence length while preserving exactness.

While FA significantly improves end-to-end throughput for long sequences, it still underutilizes the compute peak of modern accelerators.
FlashAttention-2 (FA2) attributes this gap largely to kernel-level bottlenecks---suboptimal work partitioning across thread blocks/warps, low occupancy for single-head cases, and excessive communication via shared memory \cite{dao2023flashattention}.
FA2 refines both algorithm and kernel mapping: (i) it reduces non-matmul work in the online-softmax pipeline, (ii) parallelizes attention for a single head across multiple thread blocks to raise occupancy, and (iii) improves intra-block warp specialization to cut shared-memory traffic.
These changes bring utilization closer to GEMM-like efficiency on contemporary GPUs.

FlashAttention-3 further demonstrates that attention kernels remain sensitive to evolving hardware features.
Targeting Hopper GPUs, FA3 leverages asynchronous data movement (e.g., TMA) and warp specialization to overlap memory transfers with Tensor Core computation, and interleaves block-wise matmul with softmax updates to increase pipeline efficiency \cite{shah2024flashattention}.
In addition, FA3 introduces low-precision pathways (notably FP8 with block-wise quantization) to accelerate compute while maintaining numerical accuracy, showing that correctness in low precision hinges on careful kernel scheduling and stabilization.
Overall, these works optimize exact attention by co-designing IO-aware formulations with kernel-level scheduling, parallelization, and numerical handling in the online-softmax update path.

FlashDecoding++ \cite{hong2023flashdecoding++} is designed as a decode-time LLM inference engine, where the main optimizations include asynchronized softmax with a unified maximum, flat-GEMM optimization, and heuristic dataflow selection to better utilize GPU resources under shape-dependent workloads.
By contrast, VFA is a lightweight modification to the \emph{online-softmax statistic update} in FlashAttention-style exact attention kernels. 
As a result, FlashDecoding++ mainly improves \emph{decode-system efficiency} through scheduling and synchronization design, whereas VFA improves \emph{kernel-level efficiency} by reducing vector-limited statistic computation within the online-softmax loop.

\paragraph{KV-cache-efficient attention head variants (MQA/GQA/MLA).}
Beyond improving the asymptotic complexity of attention, a complementary line of work targets the
key--value (KV) cache cost in autoregressive decoding, where storing and reading per-head $K/V$
often becomes a dominant memory and bandwidth bottleneck.
\emph{Multi-Query Attention} (MQA) reduces KV-cache size by sharing a single set of keys/values across
all query heads, enabling faster decoding with minimal architectural change \cite{shazeer1911fast}.
\emph{Grouped-Query Attention} (GQA) generalizes this idea by partitioning heads into groups that share
$K/V$ within each group, offering a smoother accuracy--efficiency trade-off between standard multi-head
attention and MQA \cite{ainslie2023gqa}.
More recently, \emph{Multi-head Latent Attention} (MLA) compresses the KV cache into a latent representation,
aiming to significantly shrink KV memory footprint while maintaining strong model quality in long-context
settings \cite{liu2024deepseek}.

Recent work on efficient attention for long-context LLMs is commonly organized into two main families:
linear attention and sparse attention \cite{sun2025efficient}.
Both aim to reduce the quadratic time/memory complexity of full softmax attention, but they differ
fundamentally in approximation strategy and kernel implications.

\paragraph{Linear attention.}
Linear attention methods replace the softmax kernel with a form that enables associative
re-ordering of computations (or kernel feature maps), reducing complexity to linear in sequence
length \cite{katharopoulos2020transformers, choromanski2020rethinking}.
Representative approaches include kernelized attention (e.g., random feature maps and positive
mappings) and recurrent/fast-weight formulations that permit streaming updates.
While linear attention provides attractive asymptotic complexity, it typically introduces modeling
approximations and may require careful numerical treatment and kernel engineering to achieve
competitive accuracy and throughput in practice.

\paragraph{Sparse attention.}
Sparse attention, in contrast, retains the exact softmax form on selected entries, but restricts the
set of key/value tokens (or blocks) that each query attends to.
This yields efficiency by reducing the number of score computations and the amount of $V$-matrix
traffic, and often aligns naturally with block/tile-based FlashAttention kernels.

\subsubsection{ Sparse Attention}
\label{subsubsec:sparse_taxonomy}

Following the survey literature, sparse attention methods can be categorized by how the sparsity pattern
is determined \cite{nawrot2025sparse}:

\paragraph{(1) Static / pattern-based sparsity.}
These methods employ pre-defined sparsity patterns that are independent of the input content,
such as sliding-window (local) attention with a small number of global tokens.
Longformer combines local windowed attention with task-motivated global attention, providing
a drop-in sparse replacement for full attention \cite{beltagy2020longformer}.
BigBird further augments local attention with random and global connections, providing theoretical
guarantees (e.g., expressivity) while enabling longer contexts \cite{zaheer2020big}.
More recently, inference-oriented pattern designs further refine static sparsity to better match
the attention mass distribution and kernel tiling behavior.
For example, MInference proposes an ``A-shape'' sparse pattern that allocates dense compute to early
``sink'' regions and a local band while sparsifying the long-range tail, enabling efficient million-token
prompt inference without requiring content-dependent routing \cite{jiang2024minference}.
Such static patterns are kernel-friendly and predictable, but may under-utilize compute on easy cases and
lack adaptivity on hard cases.

\paragraph{(2) Dynamic / content-based sparsity.}
Dynamic sparse attention selects attended tokens/blocks based on the input content, typically via
routing, clustering, or learned selection modules.
Routing Transformer introduces content-based routing via online clustering to produce sparse attention
patterns that adapt to the sequence \cite{roy2021efficient}.
Beyond clustering-style routing, recent methods increasingly adopt lightweight scoring/indexing
to retrieve a sparse set of keys for each query.
SpargeAttention, for instance, uses sparse retrieval to identify a compact set of relevant tokens/blocks
and then performs exact attention over the selected set, improving the accuracy--efficiency trade-off for
long contexts \cite{zhang2025spargeattention}.
Similarly, DeepSeek Sparse Attention (DSA) introduces a ``lightning indexer'' that computes query-to-token
index scores and retrieves top-$k$ entries for each query token, followed by attention over the selected
key-value set \cite{liu2025deepseek}.
Finally, MoBA (Mixture of Block Attention) explores a mixture-style block selection mechanism to activate
different sparse block-attention experts, offering another route to content-adaptive sparsity at block
granularity \cite{lu2025moba}.
Dynamic sparsity can improve accuracy--efficiency trade-offs by focusing computation on relevant regions,
but often complicates implementation and may incur additional overhead for selection/routing.

\paragraph{(3) Online-statistics-driven sparsity within FlashAttention.}
A particularly kernel-compatible subclass of dynamic sparsity uses online softmax statistics
already computed inside FlashAttention-style kernels (e.g., running max/normalizer) to make skipping decisions
without auxiliary proxy scoring passes \cite{alexandridis2025flash}.
This direction is attractive for long-context inference because it preserves FlashAttention's numerical
stability machinery while directly reducing the number of processed tiles. However, due to the online running mechanism, it usually can't achieve the optimal sparsity without performance degradation.BLASST proposes a drop-in sparse attention mechanism that prunes attention blocks dynamically using
only information already available in FlashAttention-style online softmax \cite{yuan2025blasst}.

These methods primarily reduce how many tiles are processed, whereas our work focuses on reducing the per-tile statistic-update overhead, making the two directions complementary.

\section{Methods}

\subsection{Algorithm Design: Vector Relieved Flash Attention}
\label{sec:VFA}
\begin{algorithm}[t]
\caption{Vector Relieved Flash Attention}
\label{alg:vector_relived_modified}
\begin{algorithmic}[1]
\Require Query blocks $\{Q_i\}_{i=1}^{T_r}$, Key blocks $\{K_j\}_{j=1}^{T_c}$,
        Value blocks $\{V_j\}_{j=1}^{T_c}$, initialization block count $T_{c_1}$
\Ensure Output blocks $\{O_i\}_{i=1}^{T_r}$

\State \textbf{/* Precompute key-block representations */} 
\For{$j = 1$ \textbf{to} $T_{c_1}$}
  \State $k^{\mathrm{repr}}_j \leftarrow \mathrm{sabsmax}(K_j)$
\EndFor
\For{$i = 1$ \textbf{to} $T_r$}
  \State Initialize $m_i^{(0)} = -\infty,\ O_i^{(0)} = 0,\ l_i^{(0)} = 0$
  \State Initialize $m_{i,\mathrm{init}} = -\infty$ \Comment{vector, one value per row of $Q_i$}

  \For{$j = 1$ \textbf{to} $T_{c_1}$}
    \State $\mathrm{score}^{\mathrm{approx}}_{ij} \leftarrow Q_i (k^{\mathrm{repr}}_j)^\top$ \Comment{vector over rows of $Q_i$}
      \State $m_{i,\mathrm{init}}$ $ \leftarrow \max(m_{i,\mathrm{init}},\ \mathrm{score}^{\mathrm{approx}}_{ij})$ \Comment{elementwise max}
  \EndFor
  \State $m_i^{(0)} \leftarrow m_{i,\mathrm{init}}$

  \For{$j \in \langle 1,\ i,\ 2,3,\ldots,T_c\rangle\ \textbf{with}\ j\neq i\ \textbf{in the tail}$}
    \State Compute $S_{ij} = Q_i K_j^{\top}$ \Comment{Attention scores}

    \If{$j = 1\ \textbf{or}\ j = i$}
      \State $\tilde{m}_i^{(j)} = \mathrm{rowmax}(S_{ij})$ \Comment{Local maximum}
      \State $m_i^{(j)} = \max\!\bigl(m_i^{(\widetilde{j-1})},\ \tilde{m}_i^{(j)}\bigr)$ \Comment{$\widetilde{j-1}$ denotes the previous \emph{iteration index} in the $j$-loop}
      \State $\tilde{P}_{ij} = \exp\!\bigl(S_{ij} - m_i^{(j)}\bigr)$ \Comment{Attention weights}
      \State $l_i^{(j)} = e^{\,m_i^{(\widetilde{j-1})} - m_i^{(j)}}\, l_i^{(\widetilde{j-1})} + \mathrm{rowsum}(\tilde{P}_{ij})$
      \State $O_i^{(j)} = e^{\,m_i^{(\widetilde{j-1})} - m_i^{(j)}}\, O_i^{(\widetilde{j-1})} + \tilde{P}_{ij} V_j$
    \Else
      \State $m_i^{(j)} = m_i^{(\widetilde{j-1})}$ \Comment{Freeze $m$ when $j \notin \{1,i\}$; $\widetilde{j-1}$ is the previous $j$-loop iteration}
      \State $\tilde{P}_{ij} = \exp\!\bigl(S_{ij} - m_i^{(j)}\bigr)$ \Comment{Attention weights}
      \State $l_i^{(j)} = l_i^{(\widetilde{j-1})} + \mathrm{rowsum}(\tilde{P}_{ij})$ \Comment{No rescale factor}
      \State $O_i^{(j)} = O_i^{(\widetilde{j-1})} + \tilde{P}_{ij} V_j$ \Comment{No rescale factor}
    \EndIf

  \EndFor
  \State $O_i = O_i^{(T_c)} / l_i^{(T_c)}$ \Comment{Final normalization}
\EndFor
\State \Return $\{O_i\}_{i=1}^{T_r}$
\end{algorithmic}
\end{algorithm}

Algorithm~\ref{alg:vector_relived_modified} presents VFA, a hardware-friendly variant of FlashAttention-style online softmax.
VFA is motivated by a practical observation: in modern attention kernels, the non-matmul components of online softmax---most notably the per-block \texttt{rowmax} reduction and the subsequent rescale chain---can become vector/SIMD limited and dominate latency when tiled matmuls already approach high efficiency.
The goal of VFA is therefore to reduce the frequency of \texttt{rowmax}-driven updates of the running maximum $m_i$ while keeping the same online-softmax state variables and kernel interface as FlashAttention.

\paragraph{Online-softmax state.}
For each query block $Q_i$, we maintain the same running statistics as FlashAttention:
a per-row running maximum $m_i^{(j)}$, a running normalizer $l_i^{(j)}$, and an output accumulator $O_i^{(j)}$.
The final output is obtained by the standard normalization $O_i = O_i^{(T_c)} / l_i^{(T_c)}$.
This alignment ensures VFA remains compatible with FlashAttention-style kernels and numerical stabilization.

\paragraph{Step 1: fast initialization of the running maximum.}
A key difference from standard FlashAttention is that VFA does not start from $m_i^{(0)}=-\infty$.
Instead, we initialize $m_i$ using a lightweight approximation that avoids forming the full score block $S_{ij}$.
Concretely, for each key block $K_j$ in a candidate set 
$j\in\{1,\ldots,T_c\}$
, we construct a block representation
$k^{\mathrm{repr}}_j=\mathrm{sabsmax}(K_j)$, where $\mathrm{sabsmax}$ selects the element with maximum absolute value per dimension while preserving its sign.
We then compute an approximate score vector
$\mathrm{score}^{\mathrm{approx}}_{ij} = Q_i (k^{\mathrm{repr}}_j)^\top$,
which yields one approximate score per row of $Q_i$.
Taking an elementwise maximum over candidates produces $m_{i,\mathrm{init}}$, and we set $m_i^{(0)} \leftarrow m_{i,\mathrm{init}}$.
Intuitively, this step provides a high-quality initial scale for exponentiation, making subsequent max updates less frequent.
Since $k^{\mathrm{repr}}_j=\mathrm{sabsmax}(K_j)$ depends only on the key block, we precompute
$\{k^{\mathrm{repr}}_j\}_{j=1}^{T_c}$ once per attention invocation and reuse them across all query blocks $Q_i$.
For decoding with a growing KV-cache, this precomputation can be performed incrementally for newly appended key blocks.
This eliminates redundant vector reductions from the inner loop and reduces the overhead of the approximation stage. The parameter $T_{c_1}$ controls the block size used for partitioning $K$ in the initialization stage, and in our implementation we set $T_{c_1}=T_c$, so that the initialization uses the same key-block granularity as the main attention loop.

\paragraph{Step 2: reordered processing with selective max updates.}
VFA next processes key/value blocks in a reordered schedule
$j\in\langle 1, i, 2,3,\ldots,T_c\rangle$ (excluding repeated $i$ in the tail).
This ordering prioritizes two blocks that empirically concentrate large attention scores (sink and local regions), so that if a true maximum is likely to be encountered, it is encountered early.
Crucially, VFA only computes the exact per-row maximum
$\tilde m_i^{(j)}=\mathrm{rowmax}(S_{ij})$ and updates $m_i^{(j)}$ when $j\in\{1,i\}$.
For these special blocks, VFA follows the standard numerically-stable online-softmax update:
$m_i^{(j)}=\max(m_i^{(\widetilde{j-1})},\tilde m_i^{(j)})$ and applies the corresponding rescale factor
$\exp(m_i^{(\widetilde{j-1})}-m_i^{(j)})$ to maintain consistency of $l_i$ and $O_i$.

\paragraph{Step 3: freezing the running maximum for the remaining blocks.}
For all other blocks $j\notin\{1,i\}$, VFA freezes the running maximum,
$m_i^{(j)}=m_i^{(\widetilde{j-1})}$, and skips the costly \texttt{rowmax} computation as well as the rescale chain.
Given a fixed $m_i^{(j)}$, VFA still computes the attention weights
$\tilde P_{ij}=\exp(S_{ij}-m_i^{(j)})$ and accumulates
$l_i^{(j)}=l_i^{(\widetilde{j-1})}+\mathrm{rowsum}(\tilde P_{ij})$ and
$O_i^{(j)}=O_i^{(\widetilde{j-1})}+\tilde P_{ij}V_j$.
This design eliminates per-block reductions and rescale-related vector operations on the majority of blocks, thereby relieving vector pressure in the kernel.

\paragraph{Discussion and implications.}
Compared with FlashAttention, VFA trades frequent max updates for a fast initialization plus selective exact updates on a small set of blocks.
Compared with threshold-based block skipping, VFA targets a complementary bottleneck: it reduces the overhead of computing the statistics (especially \texttt{rowmax}) rather than relying on these statistics to decide skipping.
As a result, VFA can be combined with dynamic sparsification methods to further reduce the number of processed blocks, while keeping the online-softmax accumulation structure unchanged.

\section{Performance Evaluation}

\begin{table}[htbp]
\centering
\caption{Comparison of FA and VFA Computation Procedures}
\label{tab:vfa_op_compare}
\begin{tabular}{|c|c|c|}
\hline
\textbf{Operation} & \textbf{FA Formula} & \textbf{VFA Formula} \\
\hline
MUL
& $S_{ij} = S_{ij}\times \mathrm{scale}$
& $S_{ij} = S_{ij}\times \mathrm{scale}$ \\
\hline
MAX
& $\tilde{m}_i^{(j)} = \mathrm{rowmax}(S_{ij})$
& - \\
\hline
MAX
& $m_i^{(j)} = \max\!\bigl(m_i^{(j-1)},\ \tilde{m}_i^{(j)}\bigr)$
& - \\
\hline
SUB
& $\bar{S}_{ij} = S_{ij} - m_i^{(j)}$
& $\bar{S}_{ij} = S_{ij} - m_i^{(j)}$ \\
\hline
EXP
& $\tilde{P}_{ij} = \exp\!\bigl(\bar{S}_{ij}\bigr)$
& $\tilde{P}_{ij} = \exp\!\bigl(\bar{S}_{ij}\bigr)$ \\
\hline
SUM
& $\hat{l}_i^{(j)} = \mathrm{rowsum}(\tilde{P}_{ij})$
& $\hat{l}_i^{(j)} = \mathrm{rowsum}(\tilde{P}_{ij})$ \\
\hline
MAD
& $l_i^{(j)} = e^{\,m_i^{(j-1)} - m_i^{(j)}}\, l_i^{(j-1)} + \hat{l}_i^{(j)}$
& $l_i^{(j)} =  l_i^{(j-1)} + \hat{l}_i^{(j)}$ \\
\hline
MUL
& $\tilde{O}_i^{(j-1)} = e^{\,m_i^{(j-1)} - m_i^{(j)}}\, O_i^{(j-1)}$
& - \\
\hline
update
& $O_i^{(j)} = \tilde{O}_i^{(j-1)} + \tilde{P}_{ij} V_j$
& $O_i^{(j)} = {O}_i^{(j-1)} + \tilde{P}_{ij} V_j$ \\
\hline
\end{tabular}
\end{table}

\begin{figure}[H]
\centering
\begin{tikzpicture}
\begin{axis}[
  ybar,
  bar width=7pt,
  width=12cm,
  height=6cm,
  enlarge x limits=0.22,
  legend style={at={(0.5,1.05)},anchor=south,legend columns=-1},
  symbolic x coords={C16V32, C8V32, C4V32, C4V16, C4V16-2xExp},
  xtick=data,
  xticklabels={C16V32, C8V32, C4V32, C4V16, {C4V16 \\ (2$\times$Exp)}},
  xticklabel style={anchor=north, align=center},
  ylabel={Latency ratio (\%)},
  ymajorgrids,
  ymin=0, ymax=100,
]
\addplot coordinates {(C16V32,100) (C8V32,50) (C4V32,17) (C4V16,17) (C4V16-2xExp,17)};
\addplot coordinates {(C16V32,46) (C8V32,46) (C4V32,46) (C4V16,46) (C4V16-2xExp,14)};
\addplot coordinates {(C16V32,77) (C8V32,77) (C4V32,85) (C4V16,43) (C4V16-2xExp,24)};
\addplot coordinates {(C16V32,46.2) (C8V32,46.2) (C4V32,54.4) (C4V16,27.3) (C4V16-2xExp,15.2)};

\legend{Tensor, Exponential, Vector (FA), Vector (VFA)}
\end{axis}
\end{tikzpicture}
\caption{Latency ratio normalized to Tensor on C16V32.}
\label{fig:vfa_latency_ratio}
\end{figure}
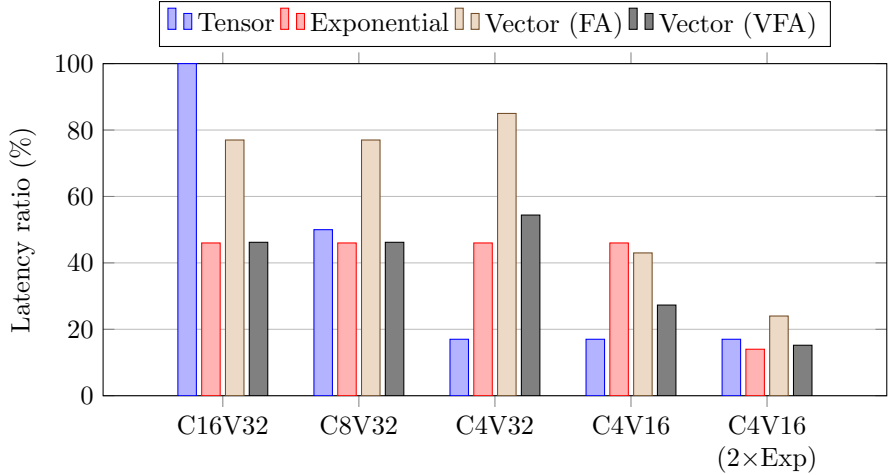

\paragraph{Implementation benefits of VFA.}
To quantify the practical benefits of VFA over standard FA, we compare their operator-level computation procedures at the granularity of a single $(Q_i,K_j,V_j)$ block interaction.
VFA introduces only a lightweight preprocessing stage. The extraction of $k^{\mathrm{repr}}_j=\mathrm{sabsmax}(K_j)$ can be naturally fused into earlier computation, and the subsequent construction of $m_{i,\mathrm{init}}$ is implemented using pure tensor operations, contributing less than \(1\%\) of the total tensor computation of standard FA.
Therefore, we focus on the differences inside the main attention accumulation loop.

Table~\ref{tab:vfa_op_compare} summarizes the per-block procedure differences between FA and VFA.
In FA, each iteration updates the running maximum via a local row-wise maximum $\tilde m_i^{(j)}=\mathrm{rowmax}(S_{ij})$ and a running $\max$ reduction, and then applies a rescaling factor $e^{m_i^{(j-1)}-m_i^{(j)}}$ to both the normalization term $l_i$ and the partial output $O_i$.
These steps introduce additional vector-linear work (MAX reductions and rescale-related MUL/MAD operations), as well as extra data dependencies and off-core traffic for rescaling the accumulated states.
In contrast, VFA first initializes the per-row maximum with an inexpensive approximation, $m_i^{(0)} \leftarrow m_{i,\mathrm{init}}$, and then freezes $m$ for the majority of blocks (i.e., $j\notin\{1,i\}$ under our schedule).
Consequently, VFA avoids repeated $\mathrm{rowmax}$ and skips the rescale multiplications on $l_i$ and $O_i$ in the update rules for those blocks.
Importantly, the tensor-dominant primitives (QK and PV) and the exponential pipeline (SUB/EXP/SUM) remain unchanged; the speedup primarily stems from reducing vector-linear operations and their associated synchronization and bandwidth overheads.

Figure~\ref{fig:vfa_latency_ratio} reports the normalized latency breakdown on the latest architecture.
On this architecture, the peak Tensor throughput for FP16, FP8, and FP4 is in the ratio of \(1\!:\!2\!:\!6\), while the exponential pipeline delivers  \(1/250\) of FP16 Tensor throughput, and the vector pipeline provides \(8\) times the exponential throughput.
Compared with FA, VFA reduces the latency contribution of the vector portion from approximately $77\%$ to $46\%$ in the C16V32 setting, and it maintains a similar ($\sim 46\%$) vector ratio under C8V32.
Under C4V32, the vector component drops from roughly $85\%$ to $54\%$, indicating that the benefit becomes more pronounced as the configuration increases the relative weight of vector-side load overhead in FA.
When moving to C4V16, VFA further decreases the vector share to about $27\%$, after which the execution becomes increasingly bounded by the exponential/SFU pipeline whose contribution remains comparatively stable.
This explains why the observed gains for C8V16 and C4V16 can become similar: once the kernel enters an exp-SFU--bound regime, further reductions in linear vector work yield diminishing end-to-end returns.

This trend also helps clarify the relation to FA4-style optimization.
FA4 reduces the softmax bottleneck mainly by approximating \(\exp\) with low-latency linear instructions, effectively converting part of the exponential cost into vector-executable work and thereby allowing the exp and vector workloads to be pooled onto the same linear pipeline.
Under such a mechanism, reducing vector latency remains beneficial even in regimes that would otherwise appear exp-bound, because the approximated exponential path itself now competes for the same linear execution resources.
From this perspective, VFA is complementary to FA4: FA4 alleviates the exponential bottleneck by remapping it to linear instructions, while VFA directly reduces the linear-side overhead caused by repeated \texttt{rowmax}-driven updates and rescale operations.
Consequently, when combined with FA4-style exp approximation, the vector-load reduction introduced by VFA can still translate into additional performance gains, including in configurations where the baseline implementation would already be limited by the exp/SFU path.

Looking forward, if the architecture approximately doubles the throughput of exp16, the exp/SFU bottleneck would be substantially relaxed.
In that scenario, the vector-side savings provided by VFA are expected to translate more directly into overall kernel speedups, and the achievable gain in configurations such as C4V16 would increase accordingly, with the limiting factors shifting back toward tensor compute and memory traffic.

\section{Experiments}

\subsection{Research Questions}
\label{subsec:rqs}

The experiments are designed to answer four questions aligned with the algorithmic components of VFA/VSA and the empirical analyses in Section~\ref{sec:data_analysis}.

\begin{itemize}
  \item \textbf{RQ1 (Accuracy preservation).}
  Does VFA preserve downstream task accuracy relative to FA2 when it freezes the running maximum on most blocks and reduces \texttt{rowmax}-driven updates?

  \item \textbf{RQ2 (Component necessity).}
  Which design elements are necessary to maintain accuracy under max-freezing?
  In particular, how do (i) $m$-initialization, (ii) sink+local reordering, and (iii) the choice of key-block representation $k^{\mathrm{repr}}_j$ affect accuracy?

  \item \textbf{RQ3 (Approximation fidelity).}
  Is the approximation stage required to be \emph{row-wise} (i.e., $\mathrm{score}^{\mathrm{approx}}_{ij}=Q_i(k^{\mathrm{repr}}_j)^\top$), or can it be replaced by a cheaper \emph{block-wise} approximation based on $q^{\mathrm{repr}}_i$ without degrading accuracy?

  \item \textbf{RQ4 (Empirical justification).}
  Do the observed attention statistics---early stabilization of the running maximum at sink/local blocks, intra-block heterogeneity of $Q/K$, and the existence of middle-block maxima---explain the success/failure modes of the ablated variants and justify the design choices of VFA?
\end{itemize}

\subsection{Experimental Setup}
\label{subsec:exp_setup}


\paragraph{Models.}
We evaluate three instruction-tuned decoder-only LLMs:
\textbf{Qwen3-30B-A3B}, \textbf{Qwen3-8B}, and \textbf{Llama3-8B}.
For all models we use the default tokenizer/chat template from the corresponding checkpoints and disable any additional sampling randomness for accuracy evaluation.

\paragraph{Baselines.}
We compare against FA2 as the primary baseline \cite{dao2022flashattention, dao2023flashattention}.
For each model and task, the ``Baseline'' column in Table~\ref{tab:acc_results_vfa_models_mean} corresponds to FA2 under the same decoding and prompt protocol as VFA.

\paragraph{Datasets and Tasks.}
We cover representative reasoning, knowledge, and coding workloads:
(i) \textbf{MATH500} for mathematical reasoning,
(ii) \textbf{MMLU} subsets (\texttt{abstract\_algebra}, \texttt{college\_computer\_science}, \texttt{\seqsplit{computer\_security}}) for English multiple-choice knowledge evaluation,
(iii) \textbf{HumanEval} for code generation,
and (iv) \textbf{CMMLU} subsets (\texttt{college\_mathematics}, \texttt{philosophy}) for Chinese multiple-choice evaluation.
All multiple-choice tasks are evaluated with exact-match accuracy under a fixed few-shot setting, and HumanEval is evaluated using standard pass@1-style greedy decoding.



\paragraph{Implementation Details.}
VFA is implemented inside the FlashAttention-2-style kernel as a conditional update rule on the running max and normalizer.
We ensure functional equivalence of non-attention components (tokenization, prompting, stopping criteria, and post-processing) across baseline and VFA.

\subsection{Accuracy Studies}
\label{subsec:acc_vfa}

We evaluate whether VFA preserves task-level accuracy relative to FlashAttention-2.
Table~\ref{tab:acc_results_vfa_models_mean} reports mean accuracy (or pass@1 for HumanEval) across three model families.

Overall, VFA matches the baseline on most tasks and shows small gains in several cases.
These results suggest that VFA's policy---reducing \texttt{rowmax}-driven online-softmax updates and freezing the running maximum on most blocks while retaining baseline updates on a small set of informative blocks---does not introduce systematic degradation in accuracy under our evaluation protocol.
The few observed regressions are small in absolute magnitude for most tasks and appear dataset/model dependent, indicating that the approximation is stable across the tested tasks/models.. 

\begin{table}[h]
\centering
\caption{Accuracy results of VFA on different models.}
\label{tab:acc_results_vfa_models_mean}
\small
\setlength{\tabcolsep}{2pt}
\renewcommand{\arraystretch}{1.12}
\begin{tabular}{T{5cm} C{1.5cm} C{1.5cm} | C{1.5cm} C{1.5cm} | C{1.5cm} C{1.5cm}}
\toprule
\multirow{2}{*}{Tasks} &
\multicolumn{2}{c|}{Qwen3-30B} &
\multicolumn{2}{c|}{Qwen3-8B} &
\multicolumn{2}{c}{Llama3-8B} \\
\cmidrule(lr){2-3}\cmidrule(lr){4-5}\cmidrule(lr){6-7}
& Baseline & VFA & Baseline & VFA & Baseline & VFA \\
\midrule
MATH500 & \nw{$0.424$} & \nw{$0.42$} & $0.25$ & $0.262$ & $0.134$ & $0.146$ \\
\seqsplit{MMLU\_abstract\_algebra} & \nw{$0.75$} & \nw{$0.76$} & $0.61$ & $0.61$ & $0.36$ & $0.35$ \\
\seqsplit{MMLU\_college\_computer\_science} & \nw{$0.75$} & \nw{$0.75$} & $0.72$ & $0.70$ & $0.45$ & $0.46$ \\
\seqsplit{MMLU\_computer\_security} & \nw{$0.81$} & \nw{$0.83$} & $0.80$ & $0.82$ & $0.82$ & $0.82$ \\
HumanEval & \nw{$0.7683$} & $0.7439$ & $0.6402$ & $0.628$ & $0.372$ & $0.372$ \\
\seqsplit{CMMLU\_college\_mathematics} & \nw{$0.619$} & \nw{$0.619$} & $0.5905$ & $0.6$ & $0.3810$ & $0.3905$ \\
CMMLU\_philosophy & \nw{$0.8571$} & \nw{$0.8476$} & $0.8286$ & $0.8190$ & $0.6286$ & $0.6190$ \\
\midrule\midrule
\textbf{Avg.} &  & \textbf{$-0.0011$} &  & \textbf{$+0.0000$}&  & \textbf{$+0.0017$}\\
\bottomrule
\end{tabular}
\end{table}

\subsection{Ablation Studies}
We conduct ablations on \texttt{MMLU\_abstract\_algebra} to isolate the contribution of the approximation stage used for $m$-initialization and the subsequent control-flow decisions (reordering and max-freezing).
Unless otherwise specified, all variants share the same  configuration, numerical precision, and evaluation protocol, so that differences are attributable to the ablated component.

\paragraph{Baseline and full method.}
\textbf{Baseline (FA2)} corresponds to the standard FlashAttention-style online softmax without our approximation stage.
\textbf{VFA (full, $\mathrm{sabsmax}$)} is the complete proposed pipeline, using the signed-absmax block representation $k^{\mathrm{repr}}_j=\mathrm{sabsmax}(K_j)$ for $m$-initialization and the reordered schedule with selective max updates on special blocks.
As shown in Table~\ref{tab:ablation_krepr_mmlu_abstract_algebra}, the full method matches or slightly improves the baseline on this task, indicating that the approximation stage and the max-freezing mechanism do not harm task accuracy under the evaluated setting.

\paragraph{Ablating the \texorpdfstring{$k^{\mathrm{repr}}_j$}{k\_repr} construction.}
We first ablate the choice of key-block representation $k^{\mathrm{repr}}_j$, which determines the approximation
$\mathrm{score}^{\mathrm{approx}}_{ij}=Q_i (k^{\mathrm{repr}}_j)^\top$ used to initialize $m_i^{(0)}$.
The objective is to test whether preserving sign information and capturing extreme values per feature dimension are critical for producing a useful upper-bound-like scale for exponentiation.

\begin{itemize}
  \item \textbf{VFA (repr = $K_{\max}$)} replaces $\mathrm{sabsmax}$ with per-dimension maximum of $K_j$.
  \item \textbf{VFA (repr = $K_{\mathrm{mean}}$)} uses per-dimension mean of $K_j$, which smooths extremes but may under-estimate the true maximum score.
  \item \textbf{VFA (repr = $|K|_{\max}$, unsigned)} uses per-dimension absolute maximum but discards sign, potentially destroying alignment information between $Q$ and $K$.
\end{itemize}

All three alternatives lead to a pronounced drop in accuracy relative to \texttt{VFA (full)}.
This suggests that the approximation stage is highly sensitive to the representational choice: preserving the signed extreme (via $\mathrm{sabsmax}$) appears necessary to produce a reliable initialization of $m_i$.
In contrast, mean-based representations are overly conservative, and unsigned absmax loses directional information, both of which can yield a poor scale for the subsequent exponentiation and accumulation. These variants collapse because the initialization becomes systematically miscalibrated, and the subsequent max-freezing prevents correction.

\paragraph{Removing \texorpdfstring{$m$-initialization}{m-init}.}
\textbf{VFA (w/o $m$-init)} disables the approximation stage and falls back to the default initialization $m_i^{(0)}=-\infty$ (or the baseline initialization used in our implementation).
This ablation verifies whether our gains stem from early acquisition of a high-quality $m_i$.
The substantial degradation indicates that $m$-initialization is not merely an optional optimization; rather, it is a necessary condition for max-freezing to remain accurate, because freezing $m$ without a good initial scale can miscalibrate the exponentiation and bias the online accumulation.

\paragraph{Ablating the reordered schedule.}
\textbf{VFA (w/o reorder)} keeps the same approximation-based initialization but removes the special-block-first schedule (i.e., it processes blocks in the default sequential order).
This ablation tests whether early processing of empirically high-mass regions (sink/local) is important for correcting the initialization and absorbing rare large scores via exact max updates.
The observed accuracy drop suggests that ordering is not a cosmetic change: prioritizing special blocks early likely increases the chance of capturing true maxima before max-freezing is applied broadly.

\paragraph{Replacing row-wise approximation with block-wise approximation on \texorpdfstring{$Q$}{Q}.}
Finally, we ablate the left operand in the approximate score computation.
The default approximation produces a \emph{row-wise} score vector,
$\mathrm{score}^{\mathrm{approx}}_{ij}=Q_i(k^{\mathrm{repr}}_j)^\top$,
yielding one approximate score per row of $Q_i$ and enabling per-row initialization of $m_i$.
We replace this with a cheaper \emph{block-wise} score
$\mathrm{score}^{\mathrm{approx}}_{ij}=q^{\mathrm{repr}}_i(k^{\mathrm{repr}}_j)^\top$,
where $q^{\mathrm{repr}}_i$ summarizes $Q_i$ (e.g., $\mathrm{absmax}(Q_i)$) and the resulting scalar score is broadcast to all rows.
This ablation evaluates whether a coarse block-level summary of $Q_i$ is sufficient for initializing a per-row maximum.
The results indicate a clear accuracy loss, implying that preserving per-row variation in $Q_i$ is important for building a reliable initialization of $m_i$; collapsing $Q_i$ to a single vector over-smooths row-specific maxima and reduces the fidelity of the approximation stage.

\paragraph{Takeaway.}
Overall, the ablation results confirm that (i) the approximation stage is a key enabler of VFA, (ii) the signed-absmax key representation is crucial for maintaining accuracy, and (iii) both the special-block-first schedule and the row-wise (rather than block-wise) approximation contribute meaningfully to the robustness of $m$-initialization under max-freezing.

\begin{table}[H]
\centering
\caption{Ablation on \texttt{MMLU\_abstract\_algebra}: block representation $k^{\mathrm{repr}}_j$}
\label{tab:ablation_krepr_mmlu_abstract_algebra}
\small
\setlength{\tabcolsep}{10pt}
\begin{tabular}{lc}
\toprule
Methods & MMLU\_abstract\_algebra \\
\midrule
Baseline (FA2) & $0.75 \pm 0.0435$ \\
VFA (full, $\mathrm{sabsmax}$) & $0.76 \pm 0.0429$ \\
VFA (repr = $K_{\max}$) & $0.23 \pm 0.0423$ \\
VFA (repr = $K_{\mathrm{mean}}$) & $0.21 \pm 0.0409$   \\
VFA (repr = $|K|_{\max}$, unsigned) &  $ 0.22 \pm 0.0416$\\
VFA (w/o $m$-init) & $0.22 \pm 0.0416$ \\
VFA (w/o reorder) & $0.24 \pm 0.0429$ \\
VFA (reprQ = $\mathrm{absmax}(Q_i)$; block-wise score) & $0.22 \pm 0.0416$ \\
VFA (reprQ = $\mathrm{sabsmax}(Q_i)$; block-wise score) & $0.22 \pm 0.0416$ \\
VFA (reprQ = $Q_{\mathrm{mean}}$; block-wise score) & $0.65 \pm 0.0479$ \\
\bottomrule
\end{tabular}
\end{table}

\subsection{Data Analysis}
\label{sec:data_analysis}
\begin{figure}[H]
  \centering
  \begin{subfigure}[t]{0.49\linewidth}
    \centering
    \includegraphics[width=\linewidth]{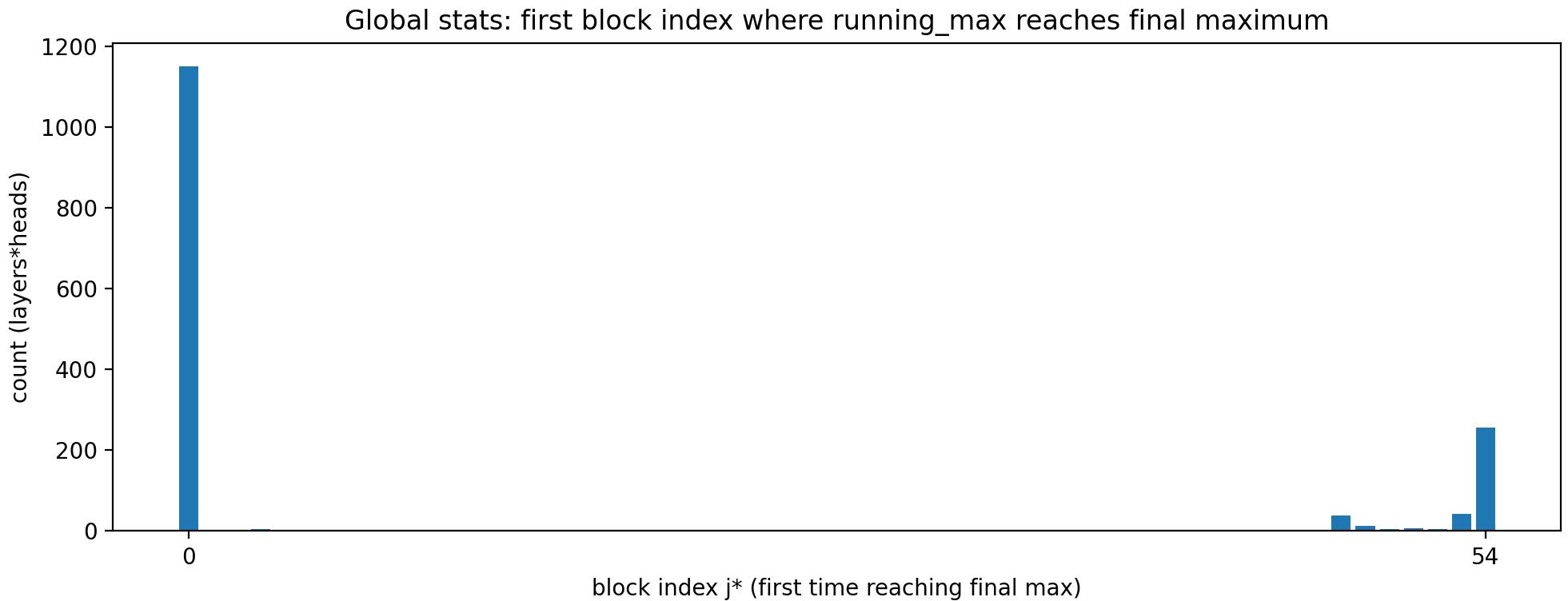}
    \caption{Running-max stabilization position $j_i^\star$.}
    \label{fig:running_max}
  \end{subfigure}
  \hfill
  \begin{subfigure}[t]{0.49\linewidth}
    \centering
    \includegraphics[width=\linewidth]{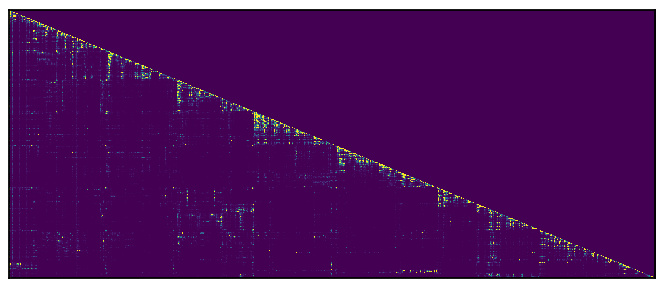}
    \caption{Attention score heatmap $P$.}
    \label{fig:p_heatmap}
  \end{subfigure}
  \caption{Empirical evidence supporting the sink+local reordering strategy.}
  \label{fig:analysis_two_figs}
\end{figure}

\begin{figure}[H]
  \centering
  \begin{subfigure}[t]{0.49\linewidth}
    \centering
    \includegraphics[width=\linewidth]{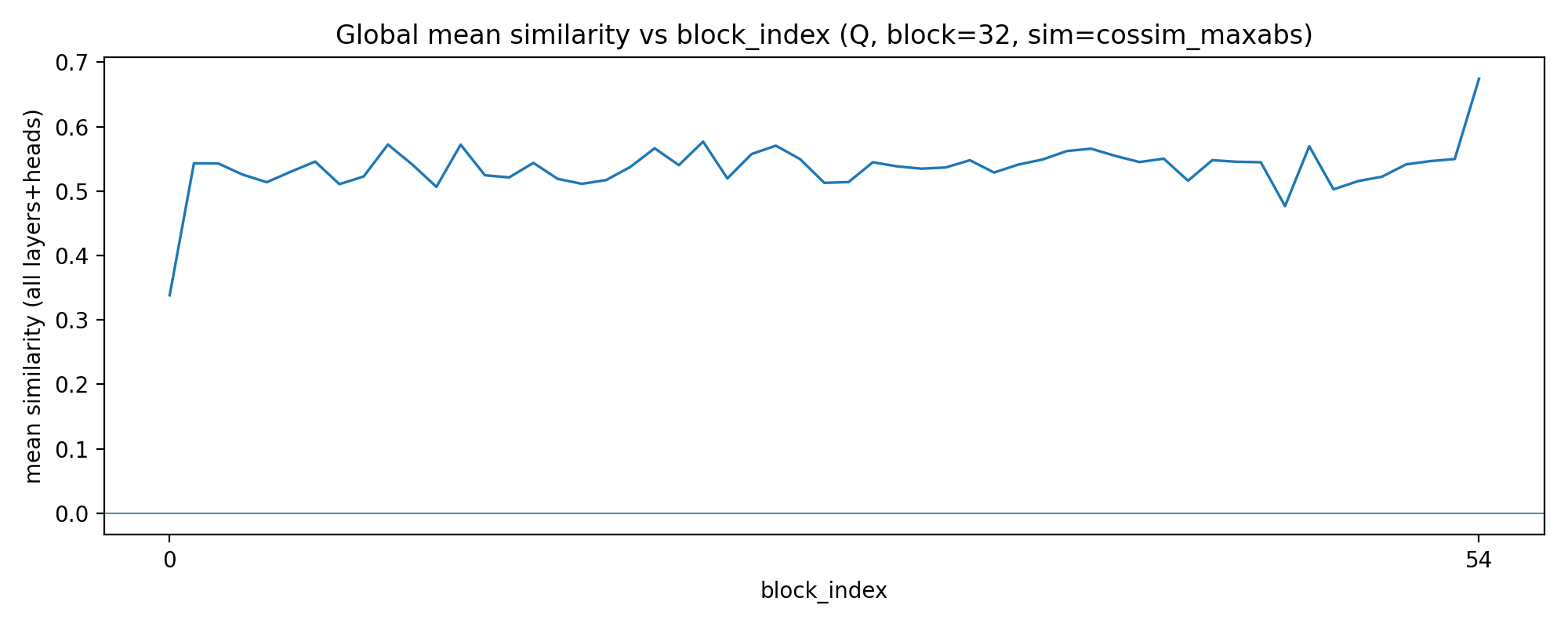}
    \caption{Block similarity of $Q$ measured by $\mathrm{cosSim}(X)$.}
    \label{fig:block_sim_q}
  \end{subfigure}
  \hfill
  \begin{subfigure}[t]{0.49\linewidth}
    \centering
    \includegraphics[width=\linewidth]{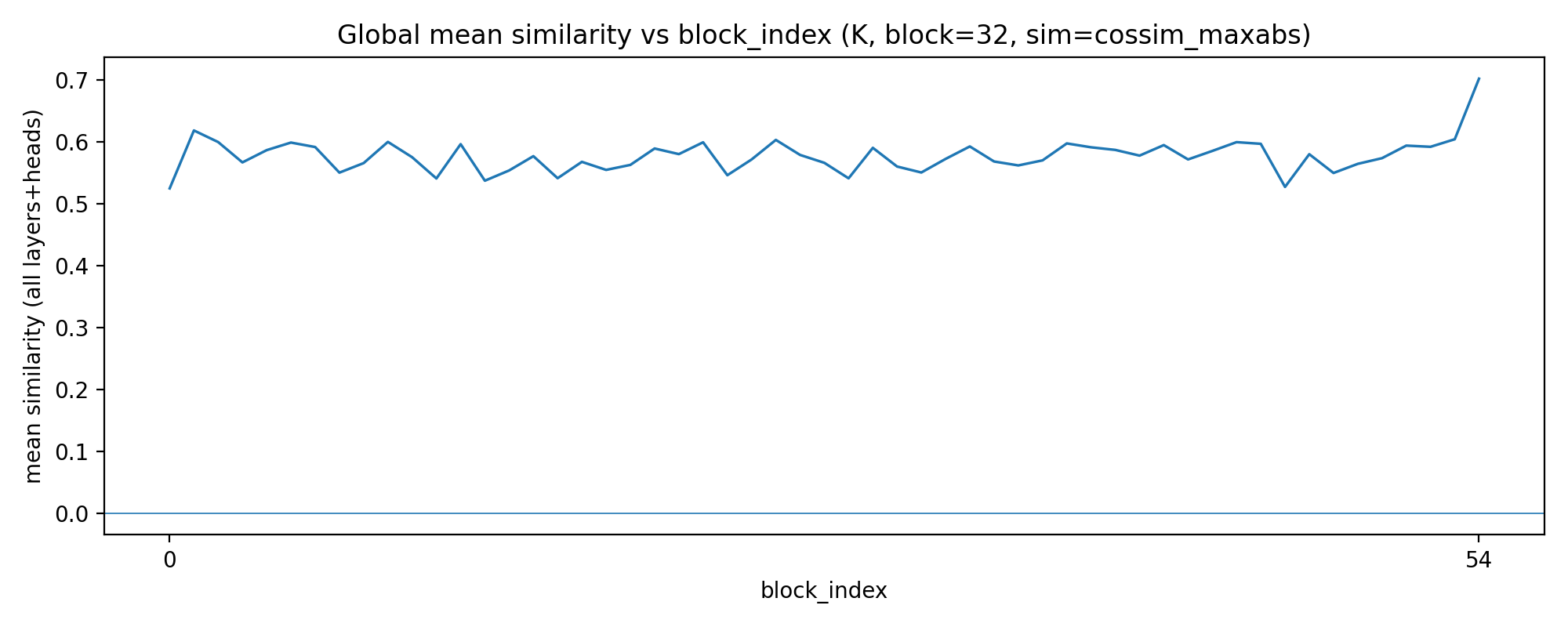}
    \caption{Block similarity of $K$ measured by $\mathrm{cosSim}(X)$.}
    \label{fig:block_sim_k}
  \end{subfigure}
  \caption{Intra-block similarity for $Q$ and $K$ blocks  using the SpargeAttention metric.}
  \label{fig:block_similarity_two_figs}
\end{figure}

\begin{figure}[H]
  \centering
  \begin{subfigure}[t]{0.49\linewidth}
    \centering
    \includegraphics[width=\linewidth]{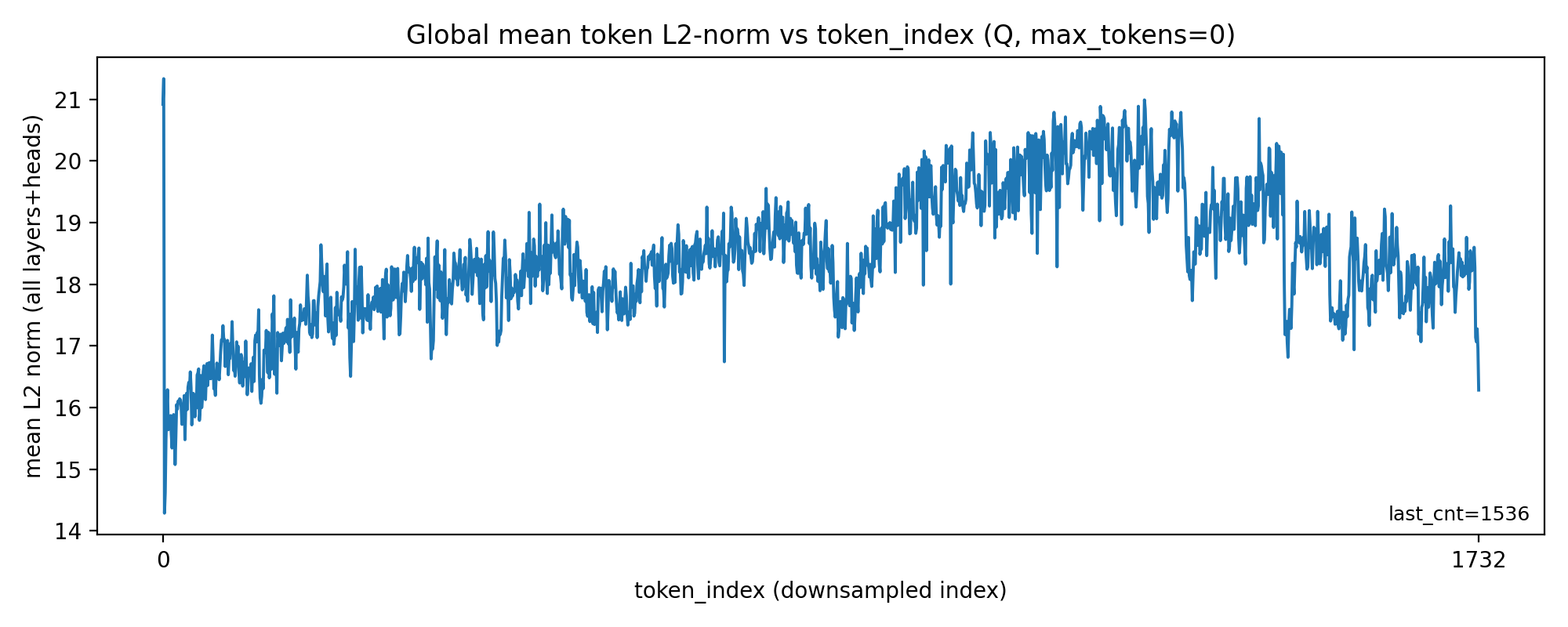}
    \caption{$\ell_2$-norm statistics of $Q$ (token/row-level).}
    \label{fig:q_l2_norm}
  \end{subfigure}
  \hfill
  \begin{subfigure}[t]{0.49\linewidth}
    \centering
    \includegraphics[width=\linewidth]{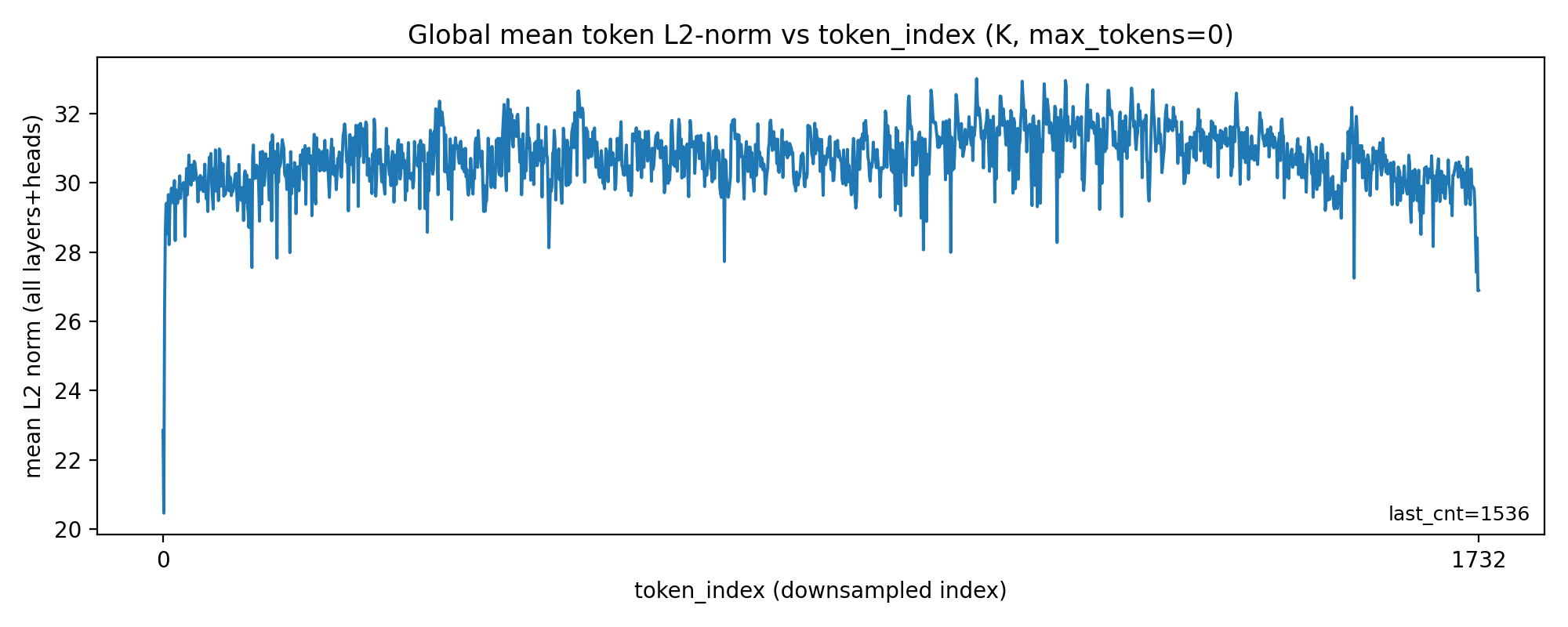}
    \caption{$\ell_2$-norm statistics of $K$ (token/row-level).}
    \label{fig:k_l2_norm}
  \end{subfigure}
  \caption{Magnitude variation within $Q$ and $K$ blocks visualized by $\ell_2$-norm statistics.}
  \label{fig:l2norm_two_figs}
\end{figure}

\subsubsection{Where the Running Maximum Stabilizes}
\label{subsec:runningmax_stabilize}

To understand why the proposed block reordering is effective, we analyze the evolution of the online-softmax running maximum during the key-block scan.
Recall that FlashAttention-style online softmax maintains a per-row running maximum $m_i^{(j)}$ as blocks are processed.
For each query block $Q_i$, we define the stabilization position as the first block index at which the running maximum reaches its final value:
\begin{equation}
j^{\star}_i \triangleq \min \Bigl\{ j \,\big|\, m_i^{(j)} = m_i^{(T_c)} \Bigr\},
\end{equation}
where the equality is evaluated elementwise for per-row maxima.
Intuitively, $j^{\star}_i$ indicates how early the true maximum score is encountered in the scan order; if $j^{\star}_i$ concentrates on a small subset of blocks, prioritizing these blocks can reduce the need for repeated max updates and rescale operations.

Figure~\ref{fig:running_max} summarizes the empirical distribution of $j^{\star}_i$.
We observe a strong concentration of stabilization events in the initial blocks and the local block (i.e., the block aligned with the query position under causal/windowed attention).
In other words, for the vast majority of query rows, the running maximum attains its final value either near the beginning of the key-block scan or when processing the local region.
This provides direct evidence that the default sequential order spends substantial effort updating $m_i$ in blocks that are unlikely to change the maximum.

\subsubsection{Attention Mass Concentration: Sink and Local Patterns}
\label{subsec:pattern_sink_local}

The above finding is consistent with the qualitative structure of attention weights.
Figure~\ref{fig:p_heatmap} visualizes the attention weight patterns (the $P$ matrix) as a heatmap.
We observe that large attention mass concentrates on a small number of regions, most prominently the early ``sink'' positions and the diagonal local neighborhood.
Such structure implies that large dot-product scores (and thus the eventual maxima that determine $m_i^{(T_c)}$) are disproportionately likely to arise from these regions, while distant off-diagonal blocks contribute comparatively small weights.

\subsubsection{Implications for Block Reordering}
\label{subsec:implication_reorder}

Together, the stabilization analysis and the heatmap patterns justify our reordering strategy that prioritizes the initial (sink) block and the local block before scanning the remaining blocks.
By bringing the blocks with the highest probability of containing the final maxima to the front of the scan, the running maximum can stabilize early, which in turn reduces the frequency of costly \texttt{rowmax} reductions and rescale updates in the online-softmax pipeline.
This empirical evidence connects the observed attention structure to the kernel-level objective of relieving vector pressure, and motivates the design choices in Algorithm~\ref{alg:vector_relived_modified}.

\subsubsection{Why Block Representations Are Ineffective}
\label{subsec:block_repr_invalid}

To explain the poor accuracy of block-representation-based variants (Table~\ref{tab:ablation_krepr_mmlu_abstract_algebra}), we perform a numerical analysis on the intra-block variability of $Q$ and $K$.
Our hypothesis is that a single vector summary (e.g., $\mathrm{max}$/$\mathrm{mean}$/$\mathrm{absmax}$) cannot faithfully represent the diverse token-level directions and magnitudes within a block, and thus yields unreliable approximations for $m$-initialization.

\paragraph{Block similarity.}
We first measure the similarity among vectors within the same block using the block-similarity metric adopted in SpargeAttention.
Given a block matrix $X$ (either a $Q$-block or a $K$-block), we define
\begin{equation}
\mathrm{cosSim}(X) \triangleq \mathrm{mean}\!\left(\frac{XX^{\top}}{\left|\max(XX^{\top})\right|}\right),
\label{eq:cosSim_sparge}
\end{equation}
where the mean is taken over all entries of the normalized Gram matrix \cite{zhang2025spargeattention}.
Figures~\ref{fig:block_sim_q} and~\ref{fig:block_sim_k} report the resulting block similarities for $Q$ and $K$, respectively.
We observe that the average block similarity is only around $0.6$, indicating that vectors within the same block are far from being well-aligned and cannot be accurately summarized by a single representative direction.

\paragraph{Norm variability.}
Direction diversity alone is not the full story: we also examine magnitude variation by plotting the $\ell_2$ norms of $Q$ and $K$ vectors.
Figures~\ref{fig:q_l2_norm} and~\ref{fig:k_l2_norm} show that the $\ell_2$ norms fluctuate substantially across tokens/rows, revealing strong intra-block scale heterogeneity.
Such large norm variation implies that extreme rows can dominate dot-product maxima, while mean- or max-based block summaries may either under-estimate or misrepresent these extremes.

\paragraph{Implication for block representations.}
Taken together, the moderate block similarity ($\approx 0.6$) and the pronounced norm fluctuations provide a direct explanation for why block representations are ineffective in our setting.
Because intra-block vectors are neither directionally coherent nor scale-homogeneous, a single block-level summary cannot preserve the information needed to approximate row-wise maxima reliably.
Consequently, block-representation-based approximations can miscalibrate $m$-initialization and lead to degraded downstream accuracy, consistent with the ablation results.

\subsubsection{Why \texorpdfstring{$m$}{m}-Initialization Is Necessary}
\label{subsec:why_m_init}

\begin{figure}[t]
  \centering
  \begin{subfigure}[t]{0.32\linewidth}
    \centering
    \includegraphics[width=\linewidth]{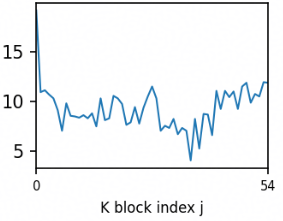}
    \caption{Block-max peaks at the sink block ($j=1$).}
    \label{fig:blockmax_sink}
  \end{subfigure}
  \hfill
  \begin{subfigure}[t]{0.32\linewidth}
    \centering
    \includegraphics[width=\linewidth]{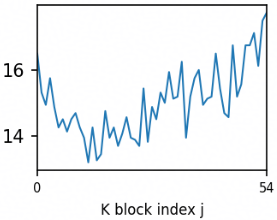}
    \caption{Block-max peaks at the local block ($j=i$).}
    \label{fig:blockmax_local}
  \end{subfigure}
  \hfill
  \begin{subfigure}[t]{0.32\linewidth}
    \centering
    \includegraphics[width=\linewidth]{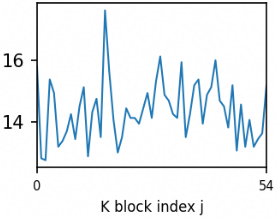}
    \caption{Block-max peaks at a middle block ($j\notin\{1,i\}$).}
    \label{fig:blockmax_middle}
  \end{subfigure}
  \caption{Representative cases of block-maximum location along the key-block index $j$, motivating the need for $m$-initialization.}
  \label{fig:blockmax_three_cases}
\end{figure}

We further investigate why $m$-initialization is necessary for the proposed max-freezing scheme.
The key challenge is that the true per-row maximum score is not always attained in the early (sink) or local blocks.
When the global maximum occurs in a middle key block, a design that relies solely on early/local exact max updates can miss the correct scale, and freezing $m$ thereafter can prevent later correction.

To quantify this phenomenon, Figures~\ref{fig:blockmax_sink}--\ref{fig:blockmax_middle} plot the block maximum as a function of the key-block index $j$ under three representative settings.
Specifically, for each query block $Q_i$ and key block $K_j$, we define the block maximum as
\begin{equation}
\tilde{m}_i^{(j)} \triangleq \mathrm{rowmax}(S_{ij}), \quad S_{ij}=Q_iK_j^{\top},
\end{equation}
and visualize how $\tilde{m}_i^{(j)}$ varies as $j$ increases.

Across the three cases, we observe a consistent pattern:
the location of the maximum block score can occur at the initial (sink) block ($j=1$), at the local block ($j=i$), and in the middle of the sequence.
While prioritizing sink and local blocks captures many maxima early, the presence of middle-block maxima implies that a strategy without $m$-initialization would systematically underestimate the true maximum scale for a non-trivial fraction of queries.
Under max-freezing, such underestimation is particularly harmful because subsequent blocks do not update $m$, and the exponentiation $\exp(S_{ij}-m_i)$ becomes miscalibrated.

These results motivate $m$-initialization as a coverage mechanism: by providing a sufficiently high initial estimate of $m_i^{(0)}$ before the main scan, VFA can remain robust even when the true maximum arises in a middle block.
In practice, $m$-initialization complements the sink+local reordering by mitigating the residual cases where maxima do not lie in the prioritized regions, which aligns with the accuracy gains observed in our ablation study.

\section{VFA Extension}
\begin{algorithm}[H]
\caption{Vector Relieved Sparse Attention}
\label{alg:vector_relived_modified_with_blasst}
\begin{algorithmic}[1]
\Require Query blocks $\{Q_i\}_{i=1}^{T_r}$, Key blocks $\{K_j\}_{j=1}^{T_c}$, 
        Value blocks $\{V_j\}_{j=1}^{T_c}$, initialization block count $T_{c_1}$, threshold $\lambda$
\Ensure Output blocks $\{O_i\}_{i=1}^{T_r}$

\State \textbf{/* Precompute key-block representations */} 
\For{$j = 1$ \textbf{to} $T_{c_1}$}
  \State $k^{\mathrm{repr}}_j \leftarrow \mathrm{sabsmax}(K_j)$
\EndFor
\For{$i = 1$ \textbf{to} $T_r$}
  \State Initialize $m_i^{(0)} = -\infty,\ O_i^{(0)} = 0,\ l_i^{(0)} = 0$
  \State Initialize $m_{i,\mathrm{init}} = -\infty$ \Comment{vector, one value per row of $Q_i$}

  \For{$j = 1$ \textbf{to} $T_{c_1}$}
    \State $\mathrm{score}^{\mathrm{approx}}_{ij} \leftarrow Q_i (k^{\mathrm{repr}}_j)^\top$ \Comment{vector over rows of $Q_i$}
    \State $m_{i,\mathrm{init}} \leftarrow \max(m_{i,\mathrm{init}},\ \mathrm{score}^{\mathrm{approx}}_{ij})$ \Comment{elementwise max}
  \EndFor
  \State $m_i^{(0)} \leftarrow m_{i,\mathrm{init}}$

  \For{$j \in \langle 1,\ i,\ 2,3,\ldots,T_c\rangle\ \textbf{with}\ j\neq i\ \textbf{in the tail}$}
    \State Compute $S_{ij} = Q_i K_j^{\top}$ \Comment{Attention scores}
    \State $\tilde{m}_i^{(j)} = \mathrm{rowmax}(S_{ij})$ \Comment{Local maximum}
    \State $m_i^{(j)} = \max\!\bigl(m_i^{(\widetilde{j-1})},\ \tilde{m}_i^{(j)}\bigr)$ \Comment{Running maximum}  \State \Comment{$\widetilde{j-1}$ denotes the previous \emph{iteration index} in the $j$-loop}
    \If{$\tilde{m}_i^{(j)} - m_i^{(j)} < \ln(\lambda)$}
      \State \textbf{continue}  \Comment{\textbf{Skip this block}}
    \EndIf
    \If{$j = 1\ \textbf{or}\ j = i$}
      \State $\tilde{P}_{ij} = \exp\!\bigl(S_{ij} - m_i^{(j)}\bigr)$ \Comment{Attention weights}
      \State $l_i^{(j)} = e^{\,m_i^{(\widetilde{j-1})} - m_i^{(j)}}\, l_i^{(\widetilde{j-1})} + \mathrm{rowsum}(\tilde{P}_{ij})$
      \State $O_i^{(j)} = e^{\,m_i^{(\widetilde{j-1})} - m_i^{(j)}}\, O_i^{(\widetilde{j-1})} + \tilde{P}_{ij} V_j$
    \Else
      \State $m_i^{(j)} = m_i^{(\widetilde{j-1})}$ \Comment{Freeze $m$ when $j \notin \{1,i\}$; $\widetilde{j-1}$ is the previous $j$-loop iteration}
      \State $\tilde{P}_{ij} = \exp\!\bigl(S_{ij} - m_i^{(j)}\bigr)$ \Comment{Attention weights}
      \State $l_i^{(j)} = l_i^{(\widetilde{j-1})} + \mathrm{rowsum}(\tilde{P}_{ij})$ \Comment{No rescale factor}
      \State $O_i^{(j)} = O_i^{(\widetilde{j-1})} + \tilde{P}_{ij} V_j$ \Comment{No rescale factor}
    \EndIf

  \EndFor
  \State $O_i = O_i^{(T_c)} / l_i^{(T_c)}$ \Comment{Final normalization}
\EndFor
\State \Return $\{O_i\}_{i=1}^{T_r}$
\end{algorithmic}
\end{algorithm}

Algorithm~\ref{alg:vector_relived_modified_with_blasst} extends VFA by incorporating dynamic block sparsification \cite{yuan2025blasst}, resulting in VSA.
While VFA reduces the per-block overhead of online softmax by eliminating most \texttt{rowmax}-driven max updates and rescale operations, it still evaluates all $T_c$ key blocks for each query block.
VSA further reduces compute and memory traffic by skipping blocks whose contributions to the softmax output are provably negligible, thereby combining two orthogonal sources of speedup:
\emph{(i) fewer blocks processed} via sparsification, and \emph{(ii) cheaper processing per block} via vector-relieved online softmax.

\paragraph{BLASST-style skip criterion.}
VSA adopts the BLASST principle of using online-softmax statistics to decide whether a block can be skipped.
For each query block $Q_i$ and key block $K_j$, we compute the score block $S_{ij}=Q_iK_j^\top$ and its per-row block maximum
$\tilde m_i^{(j)}=\mathrm{rowmax}(S_{ij})$.
Given the current running maximum $m_i^{(j)}$, the quantity $(\tilde m_i^{(j)}-m_i^{(j)})$ controls a worst-case upper bound on this block’s exponentiated contribution relative to the current scale.
If this value falls below a threshold $\ln(\lambda)$, the block is skipped:
\begin{equation}
\tilde m_i^{(j)}-m_i^{(j)} < \ln(\lambda)\ \ \Rightarrow\ \ \text{skip block }(i,j),
\end{equation}
which avoids computing $\tilde P_{ij}=\exp(S_{ij}-m_i^{(j)})$, bypasses loading the corresponding $V_j$ tile, and elides the subsequent $\tilde P_{ij}V_j$ accumulation.
This skip rule is particularly attractive in a FlashAttention-style kernel because it reuses quantities already computed for numerical stabilization and introduces minimal additional control flow.

\paragraph{How sparsification interacts with max-freezing.}
A subtlety is that VFA intentionally freezes the running maximum for most blocks to avoid repeated \texttt{rowmax} reductions and rescale.
In contrast, BLASST's skip decision depends on $\tilde m_i^{(j)}$ and the current $m_i^{(j)}$.
VSA reconciles the two by (1) using $m$-initialization to provide a strong initial scale $m_i^{(0)}$, and (2) allowing exact max updates (and rescale) only on special blocks (sink/local), while still evaluating the skip condition on all visited blocks.
As a result, VSA preserves VFA's vector-relieved update path for non-special blocks, yet can skip a large fraction of blocks whose maxima are far below the initialized/updated running maximum.
In practice, this design yields a ``best of both worlds'' behavior: if the attention mass concentrates on sink/local regions, most tail blocks are either (i) skipped outright, or (ii) accumulated with frozen $m$ without rescale.

Consequently, VSA can be viewed as a composable extension: VFA addresses the vector-limited components of online softmax, while BLASST-style sparsification reduces the number of blocks that must be processed.
Preliminary experimental results for VSA are provided in the Appendix.


\section{Conclusion and Limitation}
\label{sec:conclusion}

\paragraph{Conclusion.}
This paper revisits FlashAttention-style online softmax with hardware efficiency in mind and identifies a practical bottleneck that emerges as matmul paths approach peak efficiency: per-tile statistic updates (\texttt{rowmax}/\texttt{rowsum}) and the rescale chain can become vector/SIMD limited and dominate end-to-end latency..
To address this bottleneck without changing the attention operator, we propose VFA, which reduces the frequency of \texttt{rowmax}-driven running-maximum updates via three lightweight mechanisms: (i) a fast $m$-initialization based on key-block representations, (ii) sink+local key-block reordering to prioritize high-impact regions, and (iii) max-freezing for the remaining blocks to bypass repeated reductions and rescale operations.
We further demonstrate that VFA composes naturally with BLASST-style block skipping \cite{yuan2025blasst}, yielding VSA that combines fewer processed blocks with cheaper per-block statistic updates.
Our results confirm that VFA delivers tangible speedups in regimes where online-softmax statistic updates are a primary bottleneck, while preserving model accuracy on representative benchmarks.
Relative to the C16V32 baseline, our C8V32/C4V32/C4V16 variants achieve up to $\sim2\times$ speedup; C4V16 is currently constrained by exponent throughput, and in a hypothetical  scenario with higher exp16 capacity its speedup could increase (up to $\sim6\times$).

\paragraph{Limitations and future work.}
VFA leverages the empirical observation that the running maximum often stabilizes early around sink/local regions; workloads with highly non-local or adversarial attention distributions may reduce the effectiveness of reordering and max-freezing.
A promising direction is to design adaptive schedules that detect stabilization online and dynamically adjust the freezing strategy.

\textbf{Initialization robustness.}
Although the proposed $m$-initialization mitigates cases where the true maximum occurs in middle blocks, its approximation quality can vary across layers, heads, and tasks.
Future work includes exploring stronger yet still lightweight representations (e.g., multi-prototype or norm-aware summaries) and calibrating initialization policies per head/group.

\textbf{Evaluation scope.}
While our experiments indicate no observable quality degradation on the tested benchmarks, a broader evaluation covering more models, longer-context tasks, and additional generation settings would strengthen the conclusions.
In particular, future work should include more extensive numerical stress tests for stability under extreme sequence lengths and diverse prompt distributions.



\section*{Acknowledgments}
Disclosure of AI-assisted editing. We used an LLM-based assistant for English polishing (grammar and readability) on selected passages. The technical contributions (method, performance evaluation, experiments, and conclusions) are entirely the authors’ original work, and all edits were reviewed by the authors, who take full responsibility for the manuscript.


\bibliography{library}

\bibliographystyle{abbrv}

\section*{Appendix}

\subsection*{VSA}

\begin{table}[H]
\centering
\caption{Accuracy results on \texttt{MMLU\_computer\_security}. Each cell is reported as $\mathrm{acc}\pm\mathrm{std}$ followed by
(parentheses) sparsity; for \texttt{Blasst\_FA4} we additionally report the rescale-skip rate as
(sparsity / rescale-skip). For \texttt{Blasst\_Rowskip}, sparsity is measured row-wise.}
\label{tab:acc_results_blasst_plus_mmlu_cs_only}
\small
\setlength{\tabcolsep}{6pt}
\begin{tabular}{lc}
\toprule
Method & MMLU\_computer\_security \\
\midrule
Baseline (FA2) & $0.81 \pm 0.0394$ \\
\midrule\midrule

Blasst ($\lambda=1\mathrm{e}{-1}$) & $0.8 \pm 0.0402(32.3\%)$ \\
Blasst ($\lambda=3\mathrm{e}{-1}$) & $0.84 \pm 0.0368(44.0\%)$ \\
Blasst ($\lambda=5\mathrm{e}{-1}$) & $0.83 \pm 0.0378(49.5\%)$ \\
Blasst ($\lambda=9\mathrm{e}{-1}$) & $0.83\pm 0.0378(56.1\%)$ \\
\midrule\midrule

Blasst\_SWA ($\lambda=1\mathrm{e}{-1}$) & $0.8 \pm 0.0402(37.4\%)$ \\
Blasst\_SWA ($\lambda=3\mathrm{e}{-1}$) & $0.81 \pm 0.0394 (48.9\%)$ \\
Blasst\_SWA ($\lambda=5\mathrm{e}{-1}$) & $0.79 \pm 0.0409(54.1\%)$ \\
Blasst\_SWA ($\lambda=9\mathrm{e}{-1}$) & $0.82 \pm 0.0386 (60.5\%)$ \\
\midrule\midrule

Blasst\_FA4 ($\lambda=1\mathrm{e}{-1}$) & $0.82 \pm 0.0386 (32.1 \% / 96.8 \%)$ \\
Blasst\_FA4 ($\lambda=3\mathrm{e}{-1}$) & $0.83 \pm 0.0378 (43.4 \% / 97.0 \%)$ \\
Blasst\_FA4 ($\lambda=5\mathrm{e}{-1}$) & $0.81 \pm 0.0391 (48.5\% / 97.0 \%)$ \\
Blasst\_FA4 ($\lambda=9\mathrm{e}{-1}$) & $0.82 \pm 0.0385 (54.1 \% / 97.1\%)$ \\
\midrule\midrule

Blasst\_Rowskip ($\lambda=1\mathrm{e}{-1}$) & $0.83 \pm 0.0378 (62.3\%)$ \\
Blasst\_Rowskip ($\lambda=3\mathrm{e}{-1}$) & $0.83 \pm 0.0378 (68.1\%)$ \\
Blasst\_Rowskip ($\lambda=5\mathrm{e}{-1}$) & $0.85 \pm 0.0359 (70.5 \%)$ \\
Blasst\_Rowskip ($\lambda=9\mathrm{e}{-1}$) & $0.8 \pm 0.0402 (73.5\%)$ \\
\midrule\midrule

VSA ($\lambda=1\mathrm{e}{-3}$) & $0.84 \pm 0.0368 (16.1\%)$ \\
VSA ($\lambda=3\mathrm{e}{-3}$) & $0.84 \pm 0.0368 (31.1\%)$ \\
VSA ($\lambda=1\mathrm{e}{-2}$) & $0.84 \pm 0.0368 (52.8\%)$ \\
VSA ($\lambda=1\mathrm{e}{-1}$) & $0.78 \pm 0.0368 (84.6\%)$ \\
\bottomrule
\end{tabular}
\end{table}

\paragraph{Accuracy Results}
Table~\ref{tab:acc_results_blasst_plus_mmlu_cs_only} reports task accuracy in the form
$\mathrm{acc}\pm\mathrm{std}$, followed by efficiency-related statistics in parentheses.
Unless otherwise stated, the parenthesized number denotes the block sparsity, i.e., the fraction of
$(i,j)$ attention tiles that are skipped by a BLASST-style rule.
For methods that additionally introduce a skip-rescale mechanism (e.g., \texttt{Blasst\_FA4}),
we report two numbers as $(\text{sparsity} / \text{rescale-skip})$:
the first is the block-skip ratio; the second is the fraction of non-skipped tiles for which the
online-softmax rescale chain is bypassed (equivalently, the scale factor becomes $1$).
Concretely, block sparsity is the fraction of tiles skipped by the BLASST test, which bypasses softmax evaluation and $\tilde{P}_{ij}V_j$ accumulation.
In contrast, rescale-skip measures how often a tile is still processed but the running-maximum update is
suppressed (or the used max is kept unchanged), eliminating the vector/SIMD-heavy rescale updates on $(O_i,l_i)$.


\paragraph{\texttt{Blasst} ($\lambda$).}
Rows labeled \textbf{Blasst} correspond to the vanilla BLASST integration.
For each tile, BLASST evaluates the criterion
\begin{equation}
\tilde{m}_i^{(j)} - m_i^{(j)} < \ln(\lambda),
\end{equation}
and triggers \textbf{continue} when the inequality holds.
In this branch, the kernel bypasses $\tilde{P}_{ij}$ computation, avoids loading the corresponding $V_j$ tile,
and elides the $\tilde{P}_{ij}V_j$ accumulation.
Therefore, the percentage in parentheses for \texttt{Blasst} is exactly the empirical block sparsity
(the fraction of skipped tiles) under the given $\lambda$.
As $\lambda$ increases, the threshold becomes easier to satisfy, typically yielding higher sparsity but potentially
larger accuracy degradation, reflecting the standard accuracy--sparsity trade-off of softmax-thresholded pruning.

\paragraph{\texttt{Blasst\_SWA} ($\lambda$): sink+local reordering.}
Rows labeled \textbf{Blasst\_SWA} 
retain the same BLASST
skip rule but changes the scan order to prioritize the sink and local blocks:
\begin{equation}
j \in \langle 1,\ i,\ 2,3,\ldots,T_c\rangle \quad (j\neq i\ \text{in the tail}).
\end{equation}
Because BLASST's decision depends on the running maximum $m_i^{(j)}$, reordering can materially change sparsity:
bringing high-mass regions earlier tends to increase $m_i$ sooner, making it more likely that later tiles satisfy
$\tilde{m}_i^{(j)}-m_i^{(j)}<\ln(\lambda)$ and are skipped.
Hence, the parenthesized percentage for \texttt{Blasst\_SWA} again reports block sparsity, but under a
different scan schedule that can alter both sparsity and accuracy at the same $\lambda$.

\paragraph{\texttt{Blasst\_FA4} ($\lambda$): block skip + rescale skip.}
Rows labeled \textbf{Blasst\_FA4} implement a two-stage saving mechanism.
First, it applies the standard BLASST block-skip test and continues on tiles deemed negligible.
Second, for tiles that are not skipped, it optionally skips the rescale chain when the running maximum
changes only slightly:
\begin{equation}
m_i^{(j)} - m_i^{(j-1)} \le \tau\ln 2
\quad \Rightarrow \quad
m_{i,\mathrm{used}}^{(j)} \leftarrow m_i^{(j-1)},
\end{equation}
so that the scale factor becomes $e^{m_i^{(j-1)}-m_{i,\mathrm{used}}^{(j)}}=1$.
This removes the vector/SIMD-heavy rescaling of $(O_i,l_i)$ while still computing
$\tilde{P}_{ij}=\exp(S_{ij}-m_{i,\mathrm{used}}^{(j)})$ and accumulating $\tilde{P}_{ij}V_j$.
Accordingly, each cell for \texttt{Blasst\_FA4} reports $(\text{sparsity} / \text{rescale-skip})$:
the first number is the BLASST block sparsity; the second is the rescale-skip rate among the
non-skipped tiles.

\paragraph{\texttt{Blasst\_Rowskip} ($\lambda$): row-wise thresholding.}
Rows labeled \textbf{Blasst\_Rowskip} 
refine BLASST
from block-level decisions to row-level decisions.
Instead of skipping an entire tile when the block maximum is small, it constructs a per-row keep mask
\begin{equation}
\mathrm{row\_keep} \gets \bigl(\tilde{m}_i^{(j)} - m_i^{(j)} \ge \ln(\lambda)\bigr),
\end{equation}
and sets the scores of skipped rows to $-\infty$ (thus contributing zero mass after exponentiation), while
applying rescale only to the kept rows.
If no row is kept, the kernel continues and the entire tile is skipped.
Because the pruning granularity differs, the parenthesized percentage for \texttt{Blasst\_Rowskip} reports \emph{row-wise sparsity} (the fraction of rows suppressed by row-level thresholding), rather than tile-level sparsity.

\paragraph{\texttt{VSA} ($\lambda$): composing VFA with sparsification.}
Rows labeled \textbf{VSA} correspond to Algorithm~\ref{alg:vector_relived_modified_with_blasst}, which composes
the vector-relieved online-softmax pipeline (VFA: $m$-initialization, sink+local reordering, and max-freezing)
with BLASST-style thresholded skipping.
At a high level, VSA targets two orthogonal sources of speedup:
(i) fewer tiles processed via BLASST skipping, and
(ii) cheaper per-tile load overhead via reduced \texttt{rowmax}/rescale frequency.

\end{document}